\PassOptionsToPackage{dvipdfmx}{graphicx}
\documentclass[a4paper,fleqn]{cas-dc}

\usepackage[numbers]{natbib}

\newcommand{\secref}[1]{Sec.~\ref{#1}} 
\newcommand{\figref}[1]{Fig.~\ref{#1}}

\def\tsc#1{\csdef{#1}{\textsc{\lowercase{#1}}\xspace}}
\tsc{WGM}
\tsc{QE}
\tsc{EP}
\tsc{PMS}
\tsc{BEC}
\tsc{DE}

\begin{document}
\let\WriteBookmarks\relax
\def\floatpagepagefraction{1}
\def\textpagefraction{.001}
\shorttitle{Multi-Narrow Transformation as a Single-Model Ensemble: Boundary Conditions, Mechanisms, and Failure Modes}
\shortauthors{Tatsuhito Hasegawa and Taisei Tanaka}

\title [mode = title]{Multi-Narrow Transformation as a Single-Model Ensemble: Boundary Conditions, Mechanisms, and Failure Modes}                      
\tnotemark[1]

\tnotetext[1]{This work was supported in part by the Japan Society for the Promotion of Science (JSPS) KAKENHI Grant-in-Aid for Scientific Research (C) under Grants 23K11164.}


\author[1]{Tatsuhito Hasegawa}[type=author,
                        auid=000,bioid=1,
                        prefix=Ass. Prof.,
                        orcid=0000-0002-0768-1406]
\cormark[1]
\ead{t-hase@u-fukui.ac.jp}
\ead[url]{haselab.fuis.u-fukui.ac.jp}

\credit{Conceptualization of this study, Methodology, Formal analysis, Funding acquisition, Supervision, Writing – original draft}

\affiliation[1]{organization={Graduate School of Engineering, University of Fukui},
                addressline={Bunkyo}, 
                city={Fukui},
                postcode={9108507}, 
                state={Fukui},
                country={Japan}}

\author[1]{Taisei Tanaka}[]
\fnmark[1]

\credit{Data curation, Writing - Original draft}

\cortext[cor1]{Corresponding author}
\fntext[fn1]{Affiliation at the time of submission: University of Fukui; current affiliation: private company in Japan.}


\begin{abstract}
Single-model ensembles (SMEs) have attracted attention as a way to approximate some of the benefits of deep ensembles within a single network. However, under an approximately matched parameter budget, it remains unclear whether model capacity should be concentrated in a single wide pathway or redistributed into many narrow and independent members. We investigate this question through the Multi-Narrow (MN) transformation, which converts a baseline CNN into an SME of narrow, path-wise independent branches while approximately preserving the dominant parameter budget. We systematically compare Single-Wide and Multi-Narrow configurations across different training-data regimes, architectures, and datasets. The results show that the effectiveness of MN is strongly data-dependent: weakly partitioned or baseline-wide models are preferable in data-rich settings, whereas highly partitioned MN models consistently outperform the baseline in low-data settings. This tendency is reproduced across multiple CNN architectures and image-classification datasets, suggesting that it is not specific to a single benchmark or model family. Analysis of internal representations shows that high-MN models learn more diverse and less redundant path-wise features. In low-data regimes, this diversity is broadly utilized and improves generalization, whereas in data-rich regimes, training becomes imbalanced and prediction is dominated by a small subset of paths. These findings clarify when and why Multi-Narrow transformation is effective, and provide practical guidance for allocating model capacity between width and member multiplicity under a limited budget.

\end{abstract}




\maketitle
\section{Introduction}

The performance of deep neural networks has been driven by increases in width, depth, and the sophistication of computational graph design. In practical settings, however, abundant computation, memory, and large-scale training data are not always available. Under such constraints, a fundamental design question is how model capacity should be allocated: should it be concentrated in a single wide representation, or redistributed across multiple narrow pathways? This question is closely related to both generalization performance and practical efficiency~\cite{chirkova2020ensemble,deng2021ensemble, divide}.

This issue is particularly important in the context of single-model ensembles (SMEs), which embed multiple predictive pathways within a single network. Deep ensembles often provide strong generalization and robustness, but they require training and maintaining multiple separate models. SMEs instead aim to capture some of the benefits of ensembling within a single model by implementing multiple members or branches inside one network~\cite{easy_ensemble, laurent2023packed}. However, many existing SMEs and collaborative learning approaches rely on additional training designs, such as diversity losses, distillation losses, or cooperative objectives. As a result, it is not straightforward to isolate the effect of structural capacity allocation itself~\cite{chirkova2020ensemble,deng2021ensemble, divide}. Moreover, most existing comparisons are discussed under standard-data regimes, and it remains insufficiently understood how the trade-off between width and member multiplicity changes when training data become limited.

In this study, we address this question through the Multi-Narrow (MN) transformation, a controlled transformation that converts a baseline neural network into an SME composed of many narrow and path-wise independent members. This transformation enables a systematic comparison between Single-Wide (SW) and MN configurations under an approximately fixed parameter budget. Using this framework, we study how the preferable allocation of capacity changes across progressively reduced training-data regimes, as well as across multiple Convolutional Neural Network (CNN) architectures and image-classification datasets. We further analyze internal representation similarity and path contribution in order to understand the mechanism behind the observed regime dependence.

Our results show that the effectiveness of the MN transformation depends strongly on training-data scale. When sufficient training data are available, the baseline-wide or weakly transformed configurations achieve more stable and higher performance. In contrast, under limited-data conditions, strongly transformed MN configurations consistently provide better generalization. Furthermore, high-MN models learn path-wise diverse and less redundant internal representations, but in data-rich regimes this diversity is not fully utilized, and prediction becomes dominated by only a subset of paths. These findings suggest that, in SMEs, the preferable allocation of model capacity between width and multiplicity is not fixed, but changes depending on the data regime. They also provide practical guidance for model design under limited budgets.

The main contributions of this work are summarized as follows.
\begin{itemize}
\item We systematically compare SW and MN configurations on standard CNN architectures under an approximately fixed parameter budget, and show that their relative effectiveness depends strongly on training-data scale.
\item We show that, in low-data regimes, configurations with high MN strength are consistently advantageous across multiple CNN architectures and image-classification datasets.
\item Through representation-similarity analysis and path-contribution analysis, we show that this advantage is associated with increased path-wise diversity and the way in which that diversity is utilized, while also identifying path imbalance as a failure mode in data-rich regimes.
\end{itemize}

\section{Related Works}

\subsection{Deep Ensembles}
\label{sec:rw:deep_ensembles}

\begin{figure}
    \centering
    \includegraphics[width=1\linewidth]{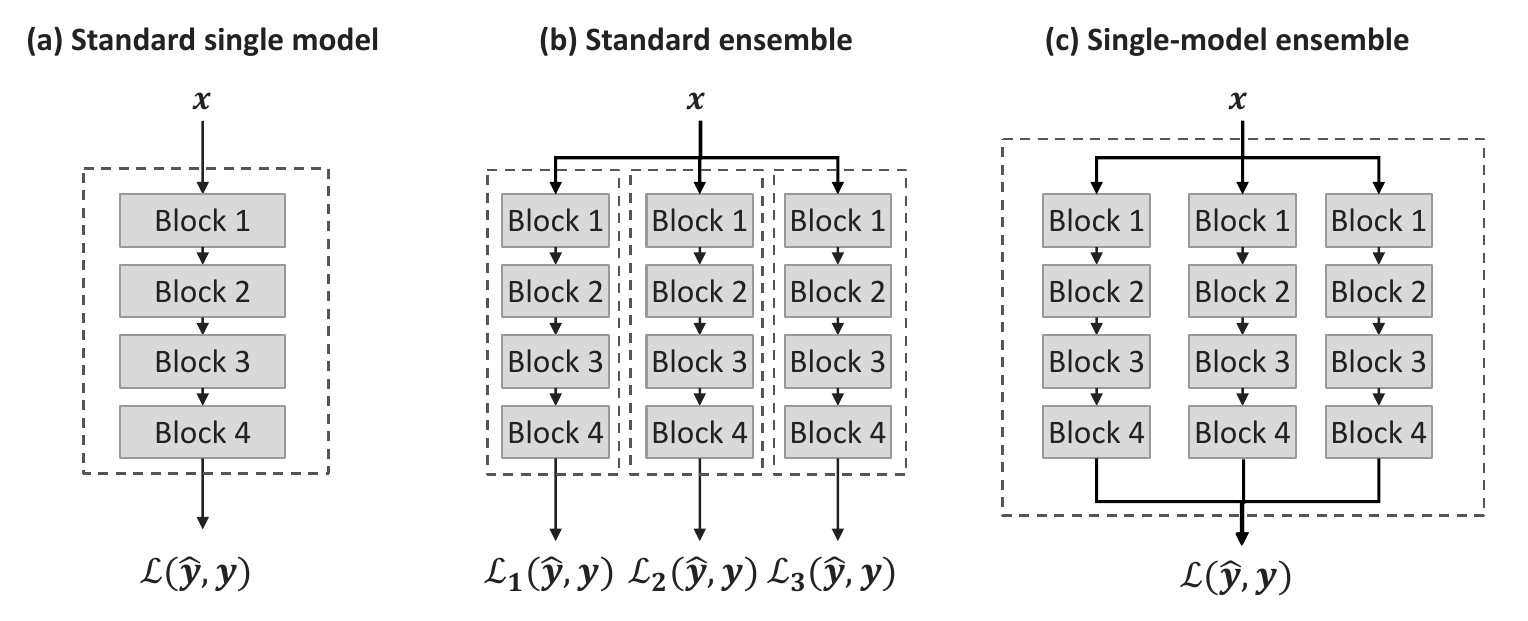}
    \caption{Three kinds of model style: standard single model, ensemble model, and SME.}
    \label{fig:style}
\end{figure}

Ensemble learning in deep learning is a framework that improves generalization performance and prediction stability over a single model by aggregating the predictions of multiple independently trained neural networks (\figref{fig:style}b). Typically, multiple models are trained with different random initializations, data orders, or bootstrap samples, and their outputs are averaged or aggregated at inference time to achieve error cancellation and variance reduction~\cite{deep_ensembles, ensemble_generalization}. In deep neural networks in particular, non-convex optimization often leads different runs to converge to different local minima, and this diversity among trained solutions is empirically known to contribute to ensemble gains~\cite{deep_ensembles}.

The main advantages of deep ensembles can be summarized as follows: (i) improved test accuracy and robustness, (ii) reduced sensitivity to training instability, and (iii) uncertainty estimation through the variance of predictive distributions. For example, in classification tasks, averaging posterior probabilities across models has been reported to mitigate overly sharp confidence estimates and improve probability calibration~\cite{Arsenii2020uncertainty,deep_ensembles}. Deep ensembles are also known to be less prone to overconfidence than single models under input perturbations and out-of-distribution (OOD) inputs.

On the other hand, the principal drawback of deep ensembles lies in their high computational cost. When $M$ models are trained independently, training computation, memory consumption, storage, and deployment cost all increase approximately by a factor of $M$, and inference also requires multiple forward passes. Therefore, in environments with strict constraints on latency, VRAM, or training cost, directly applying deep ensembles is not straightforward. This motivates the important question of how to realize ensemble-like benefits efficiently under limited computational and parameter budgets.

\subsection{SME}
\label{sec:rw:single_model_ensemble}

A standard single model (\figref{fig:style}a) consists of a single predictor trained with a single objective. In contrast, a standard ensemble (\figref{fig:style}b) trains multiple models independently and aggregates their outputs at inference time. An SME (\figref{fig:style}c) is treated externally as a single model with a single input and output, while internally containing multiple predictive pathways (ensemble members) whose outputs are aggregated within the model to obtain ensemble effects. Such a design has the advantage of keeping the loss function and training loop relatively simple compared with conventional independent ensembles, while also making it easier to analyze or prune internal members.

As concrete examples of SMEs, Coupled Ensemble~\cite{Dutt2018coupled} proposed a framework for jointly training multiple coupled models and demonstrated its effectiveness with a small number of members. Group Ensemble~\cite{chen2020groupensemblelearningensemble}, Easy Ensemble~\cite{easy_ensemble}, and Packed-Ensembles~\cite{laurent2023packed} implement multiple members efficiently within a single model using mechanisms such as group convolution. Compared with conventional ensembles that evaluate independent models sequentially, such implementations are expected to enable more efficient inference. In addition, ResNeXt~\cite{resnext} introduced the idea of \textit{cardinality}, in which many homogeneous transformations are executed in parallel at the block level and then aggregated. In this sense, it is closely related to the design philosophy of SMEs by highlighting multiplicity as a design axis distinct from width and depth.

\subsection{Collaborative Learning and Online Distillation}
\label{sec:rw:collaborative_learning}

Frameworks that jointly train multiple networks, or multiple branches within a single network, while exploiting each other's predictions or intermediate representations are often collectively referred to as collaborative learning. Unlike conventional knowledge distillation, which requires a pretrained teacher, these methods aim to improve performance by allowing learners to teach each other during training.

A representative example is Deep Mutual Learning (DML)~\cite{Zhang2018DML}, which proposes a framework in which multiple student networks are trained by mutually minimizing the KL divergence between their output distributions. In the context of online knowledge distillation, it is important to construct a high-capacity teacher signal on the fly within a single training process. ONE~\cite{Lan2018ONE} trains a multi-branch network in a single stage and reports improved performance by distilling from an online teacher obtained by aggregating the branch outputs, without requiring an additional two-stage training process. Peer Collaborative Learning (PCL)~\cite{Wu2021} further aggregates feature representations from multiple branches to construct a teacher, distills from it to each peer, and combines this process with collaborative distillation using a temporal-averaged model, thereby integrating online teacher enhancement and branch cooperation.

While these methods are highly effective, they are characterized by the introduction of additional collaborative objectives such as distillation losses or consistency losses. As a result, it is not straightforward to determine whether the observed performance improvements come from architectural capacity allocation itself or from the learning dynamics induced by the additional objectives. In contrast, the focus of this study is to understand the effect of allocating width and member multiplicity within a single model without introducing additional collaborative losses. Therefore, our work does not directly improve collaborative learning, but instead investigates the effect of structural capacity allocation from an orthogonal perspective.

\subsection{Diversity and Learner Collusion}

In ensemble learning, it has long been believed that the prediction errors of individual members should not be overly correlated; that is, predictive diversity is essential. From this viewpoint, many studies have attempted to explicitly encourage differences among members~\cite{Liu1999NCL, Chen2009NCL, Wang2010NCL, Zhang2021NCL, Zhang2024NCL}. However, diversity does not always translate into ensemble gains in a simple or monotonic manner~\cite{Danny2023Diversity}. Excessively emphasizing differences among members can sometimes harm the discriminative performance of each individual learner, implying an inherent trade-off between diversity and member strength.

In particular, when multiple members are trained under the same objective, it has been pointed out that even if multiple learners appear to exist, they may in fact converge to similar representations and predictions, leading to learner collusion~\cite{Webb2020joint, learner_collusion}. This issue is also important for SMEs, which contain multiple predictive paths within a single model. On the other hand, once auxiliary losses are introduced to explicitly optimize diversity, the loss design itself begins to affect performance, making it harder to isolate the effect of structure alone.

For this reason, our study first focuses on a single-model structure trained only with the standard discriminative loss, and analyzes how much path-wise diversity emerges naturally, under what conditions such diversity functions effectively, and under what conditions it is not fully exploited.

\subsection{Positioning of This Work}
\label{sec:positioning}

As reviewed above, existing studies on neural networks with ensemble-like structures span several distinct directions, including deep ensembles with independently trained models, memory-split ensembles that divide a fixed budget across multiple small models~\cite{chirkova2020ensemble, deng2021ensemble}, collaborative learning with auxiliary cooperative losses~\cite{Zhang2018DML,Lan2018ONE,Wu2021}, and multi-path architectures within a single model~\cite{resnext,easy_ensemble,chen2020groupensemblelearningensemble,easy_ensemble,laurent2023packed}. However, these lines of work do not necessarily share the same problem setting or research objective. Some focus on whether splitting a fixed budget can improve performance, whereas others improve learning through distillation or branch-wise cooperation.

In contrast, our interest is neither in introducing a new auxiliary objective nor in claiming a general computational advantage of multi-path architectures. Rather, we ask the following question: under approximately matched parameter budgets, should model capacity within a single model be concentrated in a single wide representation or redistributed across many narrow and independent paths, and how does the answer change depending on the low-data regime?

Table~\ref{tab:related_comparison} summarizes the position of our study relative to representative prior work. Our work is distinguished by the following combination of properties: it considers a single-model setting, compares configurations under similar parameter budgets, uses only the aggregated output for training without per-member supervision, introduces no auxiliary collaborative loss, explicitly analyzes the low-data regime, and further provides a mechanistic analysis of representation diversity and path utilization. From this perspective, our study should be viewed not as an efficiency-oriented multi-path method, but as an empirical and mechanistic investigation of regime-dependent capacity allocation in SMEs.
\begin{table*}[t]
\centering
\caption{Comparison of representative related studies and our work.
Here, ``Single model'' indicates whether multiple predictive paths or members are implemented within a single network, ``Similar budgets'' indicates whether the study explicitly compares architectures under identical or approximately matched parameter budgets, ``Aggregated-output training'' indicates whether the model is trained only through the final aggregated prediction rather than per-member supervision, ``No aux.\ loss'' indicates that no additional collaborative, distillation, or diversity-promoting loss is introduced beyond the standard supervised loss, and $\triangle$ denotes partial relevance.}
\label{tab:related_comparison}
\setlength{\tabcolsep}{4.5pt}
\renewcommand{\arraystretch}{1.15}
\begin{tabular}{lcccccc}
\toprule
Method & Single & Similar & Aggregated-output & No aux.\ & Low-data & Mechanistic \\
       & model  & budgets & training          & loss      & analysis & analysis \\
\midrule
Deep Ensembles                                  &                &                &                & $\checkmark$ &                &                \\
Chirkova et al.~\cite{chirkova2020ensemble}     &                & $\checkmark$   &                & $\checkmark$ &                &                \\
Deng and Shi~\cite{deng2021ensemble}            &                & $\checkmark$   &                & $\checkmark$ & $\triangle$    &                \\
CL~\cite{Zhang2018DML,Lan2018ONE,Wu2021}        & $\triangle$    &                &                &                &                &                \\
ResNeXt~\cite{resnext}                          & $\checkmark$   & $\triangle$    &                & $\checkmark$ &                &                \\
Easy Ensemble~\cite{easy_ensemble}              & $\checkmark$   & $\checkmark$   & $\checkmark$   & $\checkmark$ &                &                \\
DaC~\cite{divide}                               & $\checkmark$   & $\checkmark$   &                &                &                &                \\
Ours                                            & $\checkmark$   & $\checkmark$   & $\checkmark$   & $\checkmark$ & $\checkmark$   & $\checkmark$   \\
\bottomrule
\end{tabular}
\end{table*}

\section{MN Transformation}
\label{sec:multi_narrow}

In this study, we refer to the baseline model before MN transformation as the SW model. The MN transformation is a structural transformation that reconfigures the SW model into an SME with many narrow and independent predictive paths, while preserving the original input--output interface. Its purpose is to make it possible to compare a single wide representation with a representation distributed across many narrow paths under a similar parameter scale. In what follows, we define the transformation, describe its approximate parameter-preservation property in dense-coupling layers, and clarify its architectural exceptions.

\subsection{Definition of the Transformation}
\label{sec:mn_basic}

Let \(f_{\mathrm{base}}\) denote the baseline model, and let \(r \in \mathbb{N},\, r \geq 1\) be an integer hyperparameter controlling the transformation strength. The MN transformation is defined as the structural transformation
\begin{equation}
    f_{\mathrm{MN}} = \mathcal{T}(f_{\mathrm{base}}, r).
\end{equation}
Conceptually, this transformation consists of the following two operations:

\begin{enumerate}
    \item \textbf{Narrowing}: The channel width (or feature dimension) of each predictive path is reduced by a factor of \(1/r\).
    \item \textbf{Multiplication}: The narrowed subnetworks are replicated and arranged in parallel.
\end{enumerate}

In this study, the number of parallel paths is set to
\begin{equation}
    M = r^2
\end{equation}
so that the total number of parameters in the dominant intermediate layers is approximately preserved. This is because, in layers where both the input and output dimensions are reduced by a factor of \(1/r\), the number of parameters per path decreases approximately to \(1/r^2\) of the baseline. Therefore, by placing \(r^2\) such paths in parallel, the total parameter count of the dominant dense-coupling layers can be approximately preserved.

The transformed model is designed to satisfy the following properties:
\begin{enumerate}
    \item \textbf{Path-wise separation}: Except for the shared input and the final output aggregation, the paths do not share intermediate representations or parameters (\figref{fig:style}c).
    \item \textbf{I/O compatibility}: The input and output interface of the task is preserved, hence the transformed model can be used as a drop-in replacement for the baseline model.
    \item \textbf{Approximate budget preservation}: The total number of parameters is approximately preserved in the dominant intermediate layers, although exact preservation at the whole-model level depends on the architecture.
\end{enumerate}

All paths receive the same input, and their final predictions \(\mathbf{y}_m \; (m=1,\dots,M)\) are averaged to produce a single aggregated output
\begin{equation}
    \mathbf{Y} = \frac{1}{M}\sum_{m=1}^{M}\mathbf{y}_m.
\end{equation}
In this study, no path receives individual supervision; instead, the standard classification loss is applied only to the aggregated output \(\mathbf{Y}\). Thus, although the MN model is internally a multi-path structure, it is externally treated as a standard classifier with a single input and a single output.

\subsection{Parameter Preservation in Dense-Coupling Layers}
\label{sec:mn_preservation}

A central property of the MN transformation is that it preserves the total number of parameters in intermediate dense-coupling layers. Here we explain this using a convolutional layer, although the same argument also applies to fully connected layers.

Consider an intermediate convolutional layer with input channels \(C_{\mathrm{in}}\), output channels \(C_{\mathrm{out}}\), and kernel size \(K \times K\). The number of parameters in the baseline layer is
\begin{equation}
    P_{\mathrm{base}} = K^2 C_{\mathrm{in}} C_{\mathrm{out}}.
\end{equation}
Under the MN transformation, both the input and output channel dimensions of each path are reduced by a factor of \(1/r\), hence the number of parameters per path becomes
\begin{equation}
    P_{\mathrm{path}}
    = K^2 \frac{C_{\mathrm{in}}}{r} \frac{C_{\mathrm{out}}}{r}
    = \frac{1}{r^2} P_{\mathrm{base}}.
\end{equation}
If \(M=r^2\) such paths are arranged in parallel, the total number of parameters after transformation is
\begin{equation}
    P_{\mathrm{MN}}
    = M \cdot P_{\mathrm{path}}
    = r^2 \cdot \frac{1}{r^2} P_{\mathrm{base}}
    = P_{\mathrm{base}}.
\end{equation}
Therefore, the total parameter count is preserved in intermediate convolutional layers. The same reasoning applies to intermediate fully connected layers: by reducing both the input and output dimensions of each sublayer by a factor of \(1/r\) and placing \(r^2\) such sublayers in parallel, the total number of parameters is preserved.

In this sense, the MN transformation should not be interpreted as a simple model expansion, but rather as a structural transformation that reallocates the dense, highly coupled representation of SW into the sparse, weakly coupled representation of MN.

\subsection{Architectural Exceptions and Scope}
\label{sec:mn_exceptions}

The preservation rule described above mainly applies to intermediate convolutional and fully connected layers. In contrast, the same argument does not directly apply to input layers, classifiers, normalization layers, or architectures that heavily rely on depthwise-separable convolutions.

First, in the input layer, the number of input channels is fixed by the data format, and in the classifier, the output dimension is fixed by the task. Therefore, only one side of the layer can be narrowed. In such cases, the reduction per path is approximately \(1/r\) rather than \(1/r^2\), hence replicating \(r^2\) paths can increase the total number of parameters. In addition, normalization layers must be applied separately to preserve path independence, which may introduce a small increase as the total number of channels grows.

Further care is required for architectures built around depthwise-separable convolutions. While the preservation rule applies to pointwise convolutions, depthwise convolutions do not couple channels, and thus narrowing reduces their parameter count only by a factor of \(1/r\). As a result, in depthwise-heavy models such as MobileNet and EfficientNet, the increase in parameter count and theoretical computation after MN transformation may become non-negligible. Therefore, the MN transformation is most naturally suited to architectures whose dominant computation arises from standard convolutions or fully connected layers.

\subsection{Computational and Memory Implications}
\label{sec:mn_cost}

As described above, the MN transformation is designed to preserve the number of parameters in intermediate dense-coupling layers, and in architectures dominated by such layers, the theoretical computational scale also remains of the same order. However, due to the architectural exceptions discussed in the previous subsection, the total number of parameters and the total amount of computation at the whole-model level remain architecture-dependent.

Moreover, when we say that the computational cost is of the same order, we refer mainly to the theoretical operation scale derived from layer-wise parameterization, rather than to measured latency or training time. In practice, although the width of each path is reduced by a factor of \(1/r\), the number of paths increases to \(r^2\), hence the total number of intermediate feature channels retained across all paths becomes
\begin{equation}
    r^2 \cdot \frac{C}{r} = rC,
\end{equation}
which implies that the memory required for storing intermediate activations can increase. In addition, because the computation is split into many small and independent units, practical execution efficiency on GPUs may degrade due to kernel-launch overhead and fragmented memory access, even when the theoretical computation is similar.

Therefore, the cost implications of the MN transformation should be interpreted not only in terms of parameter count and theoretical computation, but also in terms of activation memory and effective hardware efficiency. In this study, we regard these as important implementation considerations, while focusing primarily on how capacity allocation under a limited budget affects predictive performance.

\section{Experiments}
\label{sec:experiments}

We evaluate the MN transformation from three perspectives: the data-regime dependence of its effectiveness, the mechanism underlying its gains in low-data regimes, and its computational implications.

\subsection{Experimental Setup}
\label{sec:exp_setup}

Unless otherwise stated, the following setup was used throughout the experiments.
We employed ResNet-18~\cite{resnet} as the baseline architecture and applied the MN transformation defined in \secref{sec:mn_basic}.
Following Easy Ensemble~\cite{easy_ensemble}, the implementation was based on group convolution, and the transformation strength was controlled by
\[
r \in \{1, 2, 4, 8, 16, 32\}.
\]
Here, \(r=1\) corresponds to the untransformed SW baseline, and \(r=32\) corresponds to a model containing \(1{,}024\) internal paths.

We used CIFAR-100 as the primary dataset.
To examine the effect of training-data scale, training samples were randomly subsampled while preserving class balance, so that the number of images per class (IPC) satisfied
\[
\mathrm{IPC} \in \{500, 200, 100, 50, 20, 10, 5, 1\}.
\]

For optimization, we used AdamW~\cite{loshchilov2018adamw} with cosine annealing~\cite{loshchilov2017cosine} as the learning-rate scheduler.
The maximum learning rate was set to
\[
\mathrm{max\_lr} = 5\times10^{-3}\times \frac{\mathrm{batch\_size}}{128},
\]
and the weight decay was fixed at \(5\times10^{-2}\).
The batch size was set to 128 by default, but adjusted in practice according to the memory usage of each model.
When the batch size was changed, the maximum learning rate was scaled proportionally.
This setting was chosen based on the common learning-rate scaling heuristic used in large-batch training~\cite{goyal2017accurate,smith2018dont}, together with preliminary experiments on CIFAR-100.

The number of training epochs was adjusted to ensure a sufficient number of parameter updates under low-data conditions:
\[
\mathrm{epochs} =
\begin{cases}
200, & \mathrm{IPC} \geq 100, \\
200 \times \left(100 / \mathrm{IPC}\right), & \mathrm{IPC} < 100.
\end{cases}
\]
That is, models were trained for 200 epochs when \(\mathrm{IPC} \geq 100\), and for proportionally longer when the data became scarcer.
In all cases, we confirmed that training had converged.

As described in \secref{sec:mn_basic}, our comparison is based on the condition that the number of parameters is approximately preserved in the dominant intermediate layers.
However, exact budget matching does not hold for input layers, classifiers, normalization layers, and architectures that heavily rely on depthwise-separable convolutions.
Accordingly, the experiments in this section should be interpreted as comparisons under approximately matched parameter budgets.

\subsection{Research Questions}
\label{sec:rqs}

We address the following two research questions:

\begin{itemize}
    \item \textbf{RQ1:} Under what training-data regimes is Multi-Narrow preferable to Single-Wide under an approximately matched parameter budget?
    \item \textbf{RQ2:} Why does high MN strength become advantageous in low-data regimes, and why does this advantage weaken or disappear in data-rich regimes?
\end{itemize}

To answer RQ1, we first conduct a controlled study on CIFAR-100 with ResNet-18 to identify the boundary conditions under which MN is beneficial, and then examine the robustness of the observed trend across architectures and datasets.
To answer RQ2, we analyze the mechanism behind the gains of high-MN models from the perspectives of representation diversity and path utilization.

\subsection{RQ1: Boundary conditions of Multi-Narrow}
\label{sec:rq1}

\subsubsection{Data-regime dependence on CIFAR-100 / ResNet-18}

\begin{figure}
    \centering
    \includegraphics[width=1\linewidth]{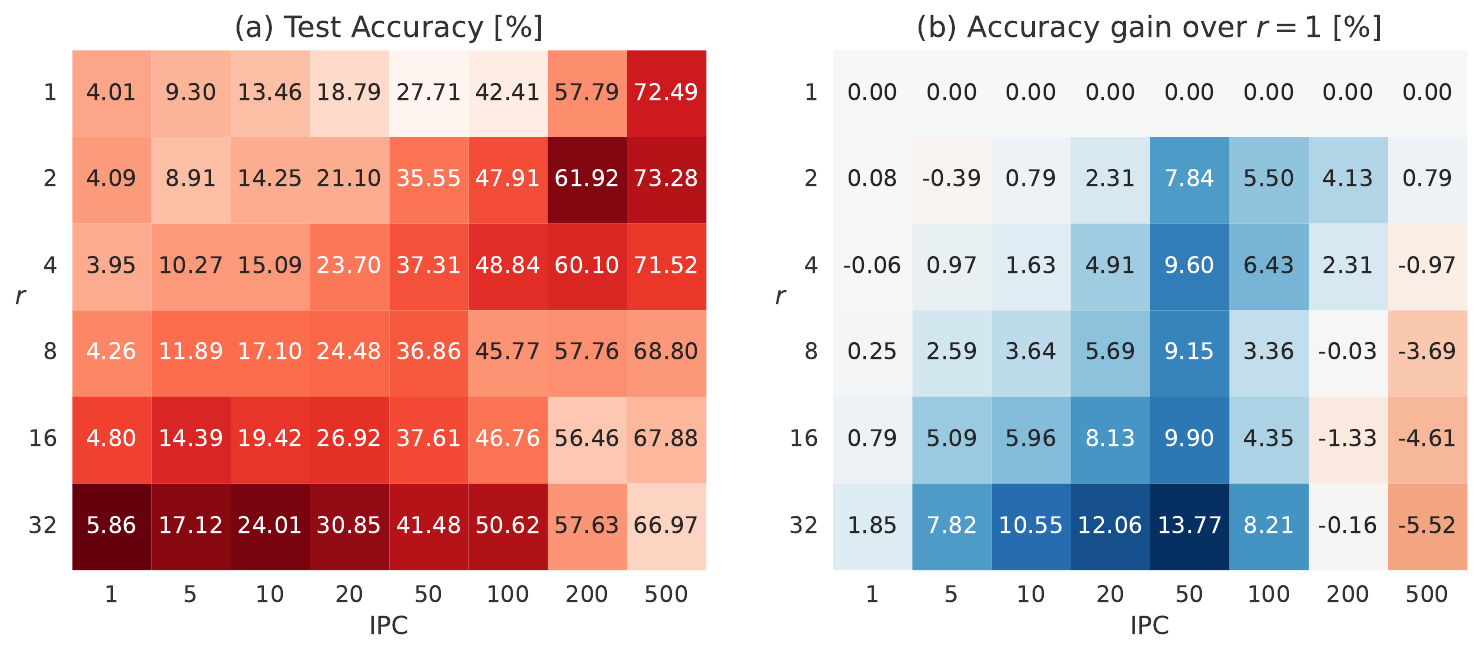}
    \caption{Effect of MN strength and training-data size on CIFAR-100 with ResNet-18.
    (a) Test accuracy, where the color intensity is normalized within each column.
    (b) Accuracy gain relative to the SW baseline (\(r=1\)).}
    \label{fig:resnet}
\end{figure}

\figref{fig:resnet} shows the results obtained by applying the MN transformation to ResNet-18 while varying both the MN strength \(r\) and the amount of training data on CIFAR-100.
The figure shows that the optimal MN strength depends strongly on the amount of training data.
When IPC is large, the baseline or weakly transformed MN models achieve higher performance, whereas stronger MN configurations become increasingly advantageous as IPC decreases.
In particular, the gains of high-MN become prominent when IPC is 50 or below, and \(r=32\) achieves the best performance gains at IPC=50.
Thus, the preferable capacity allocation is not fixed: as training data become scarcer, the preferable design shifts from SW to MN.

These results indicate that the effectiveness of MN strength is not uniform, but exhibits a clear dependence on the data regime.
When sufficient training data are available, SW or weakly transformed MN models tend to achieve higher and more stable accuracy.
By contrast, under limited-data conditions, high-MN configurations that distribute capacity across many narrow and independent paths become preferable.

\subsubsection{Optimization robustness under low-data settings}
\begin{figure}
    \centering
    \includegraphics[width=0.9\linewidth]{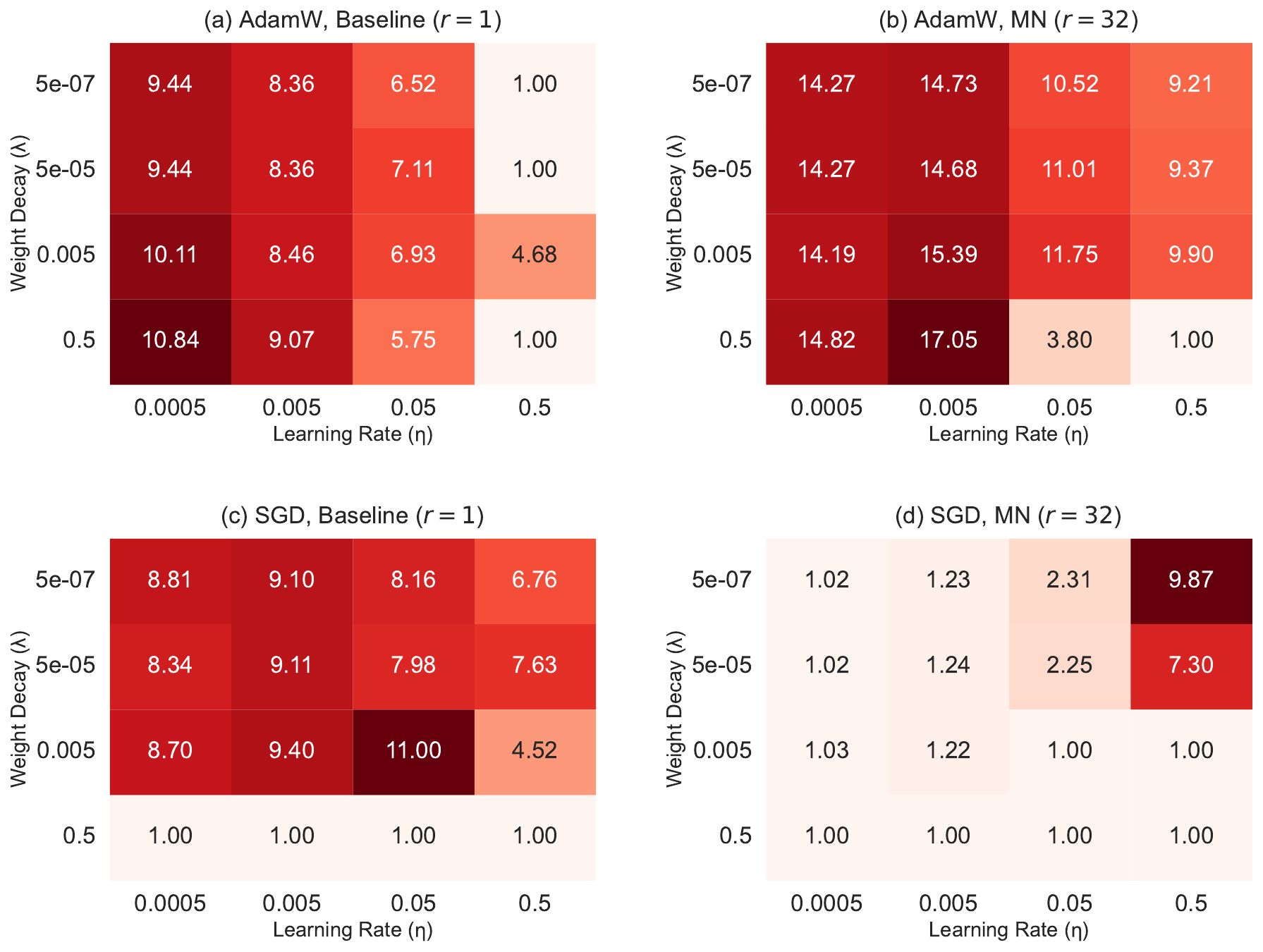}
    \caption{Supplementary hyperparameter search at IPC=5 on CIFAR-100 / ResNet-18.
    The observed advantage of high-MN is not explained solely by an obviously poor optimization setting for SW.}
    \label{fig:tuning}
\end{figure}

The results in Fig.~\ref{fig:resnet} were obtained under a common AdamW setting and may therefore not be individually optimal for the SW baseline.
To verify that the observed advantage of MN in low-data settings is not merely due to an obviously unfavorable optimization setting for SW, we conducted a supplementary hyperparameter search at IPC=5 on CIFAR-100 with ResNet-18, varying the optimizer, learning rate, and weight decay for both SW (\(r=1\)) and high MN (\(r=32\)).
The results are shown in \figref{fig:tuning}.

Although the best SW performance improved under both AdamW and SGD, it still did not reach the best performance of the high-MN model. At the same time, the high-MN model was optimized much more effectively by AdamW than by SGD, indicating that its optimization characteristics differ from those of the SW baseline. Therefore, the advantage of MN observed in the low-data regime is unlikely to be explained solely by an obviously unfavorable hyperparameter setting for SW. Rather, the supplementary results suggest that SW and high-MN differ not only in capacity allocation but also in optimization sensitivity.

\subsubsection{Robustness across architectures and datasets}

\begin{figure*}
    \centering
    \includegraphics[width=1\linewidth]{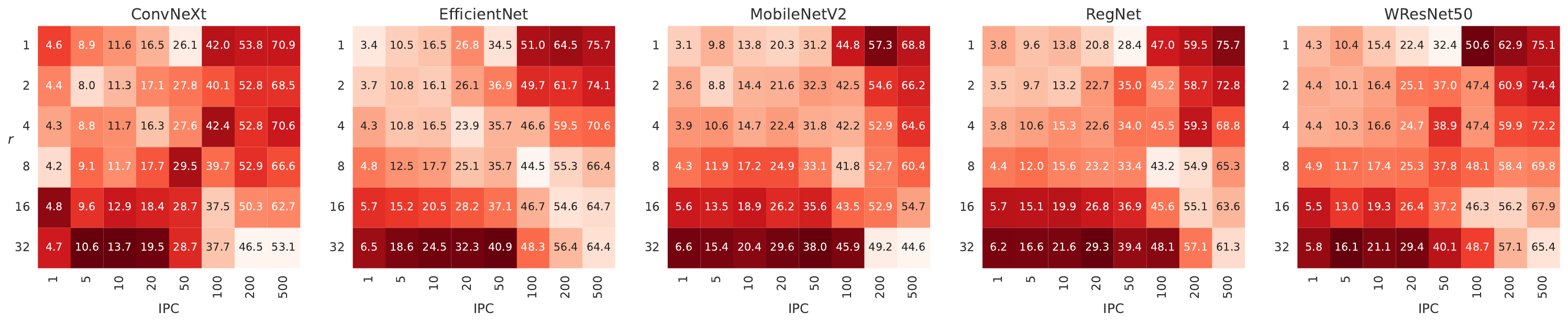}
    \caption{Robustness of the observed trend across CNN architectures.
    The preferred MN strength shifts toward larger values as the training-data regime becomes smaller across multiple architectures.}
    \label{fig:architecture}
\end{figure*}

We next examine whether the observed data-regime dependence is specific to a particular model or dataset.
First, \figref{fig:architecture} shows the robustness of the trend across architectures, including ConvNeXt-Tiny~\cite{convnext}, EfficientNet-B0~\cite{efficientnet}, MobileNetV2~\cite{mobilenet_v2}, RegNetY-400MF~\cite{regnet_y}, and Wide ResNet-50~\cite{wide_resnet}.

Although the exact transition point varies across models, many architectures exhibit the same overall tendency: higher MN strength becomes preferable as the amount of training data decreases.
In particular, below IPC=100, high-MN configurations are often the best-performing or among the best-performing models.
Thus, the advantage of high MN in low-data conditions is not specific to ResNet-18, but is reproducible across multiple CNN families.

However, as discussed in \secref{sec:mn_exceptions}, exact parameter preservation no longer holds in architectures that heavily rely on depthwise-separable convolutions, such as EfficientNet and MobileNetV2.
Accordingly, these results should be interpreted not as comparisons under strictly identical budgets, but rather as comparisons under approximately matched budgets.

\begin{figure}
    \centering
    \includegraphics[width=1\linewidth]{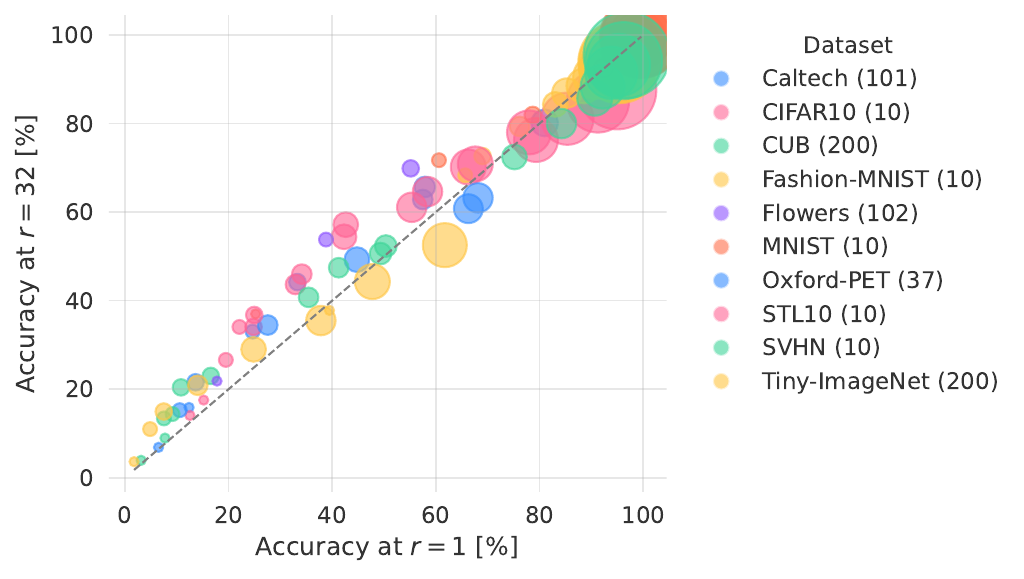}
    \caption{Robustness across datasets.
    The high-MN model (\(r=32\)) tends to outperform the SW baseline (\(r=1\)) especially in low-data conditions.}
    \label{fig:dataset}
\end{figure}

We further examine robustness across datasets in \figref{fig:dataset}.
This figure plots the test accuracy of the SW baseline (\(r=1\)) on the horizontal axis and that of the high-MN model (\(r=32\)) on the vertical axis, with bubble size proportional to the square root of the number of training samples.
In many conditions, the points lie above the diagonal, indicating that the high-MN model outperforms the baseline, especially under low-data conditions where the bubbles are smaller.
At the same time, the transition point at which MN becomes advantageous varies across datasets, suggesting that factors such as the number of classes and task difficulty influence how much data are needed before high-MN becomes beneficial.
Detailed results for each dataset are provided in the appendix.

Overall, the answer to RQ1 is that the effectiveness of the MN transformation depends strongly on the training-data regime. When sufficient data are available, SW or weakly transformed MN models are preferable, whereas in the low-data regime, high MN strength provides a reproducible advantage across multiple architectures and datasets. These observations also clarify the relation between our findings and prior work. Easy Ensemble~\cite{easy_ensemble} suggested that highly partitioned configurations can be beneficial, but its evidence was limited to 1D CNNs for activity recognition. DaC~\cite{divide} considered image classification, yet mainly focused on relatively weak partitioning and did not analyze the strongly partitioned regime under low-data conditions. Our results extend these lines of work by showing that, under approximately matched budgets, the preferred capacity allocation shifts toward stronger multiplicity as the training-data regime becomes smaller.

\subsection{RQ2: Why does high MN help in low-data regimes?}
\label{sec:rq2}

RQ1 established that high MN strength is advantageous in the low-data regime, whereas its advantage weakens in data-rich settings.
We now investigate the underlying mechanism from two perspectives:
(i) path-wise feature diversity and
(ii) the extent to which this diversity is actually utilized.

\subsubsection{Feature diversity}

\begin{figure}
    \centering
    \includegraphics[width=\linewidth]{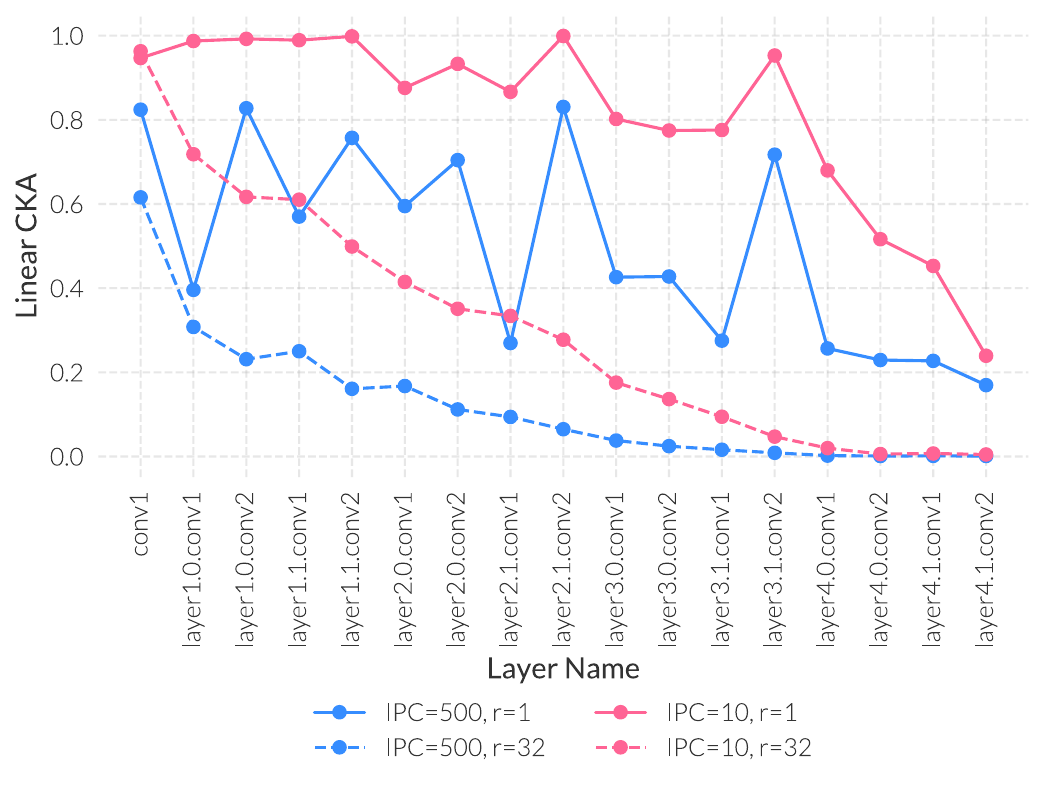}
    \caption{Layer-wise linear CKA between groups.
    High-MN models maintain more diverse path-wise representations, especially in deeper layers.}
    \label{fig:layerwise_cka}
\end{figure}

We first examine what kind of path-wise representations are formed by the MN structure.
\figref{fig:layerwise_cka} shows the linear CKA~\cite{linear_cka} similarity between groups at each layer for the SW baseline (\(r=1\)) and a high-MN model (\(r=32\)).
In the SW model, the inter-group CKA remains relatively high even in deeper layers, whereas in the high-MN model it decreases substantially toward deeper layers.
This indicates that the MN structure suppresses information sharing across paths and maintains more diverse and less redundant internal representations.

To examine whether this diversity translates into complementary predictions, we also evaluate oracle accuracy.
Here, oracle accuracy is defined as the auxiliary metric that counts a sample as correct if at least one of the \(M\) paths contained in the MN model predicts it correctly.
As shown in Table~\ref{tab:oracle_accuracy}, the oracle accuracy of high-MN models is dramatically higher than their ordinary test accuracy.
This suggests that the paths are not merely repeating the same errors, but instead are correctly handling different subsets of samples in a complementary manner.
Therefore, high MN strength can be interpreted as forming a structure that preserves multiple diverse hypotheses internally, which may serve as the basis for improved generalization under low-data conditions.

\begin{table}[tb]
  \caption{Test accuracy and oracle accuracy under different data regimes.
  The large gap in MN models indicates strong path-wise complementarity.}
  \label{tab:oracle_accuracy}
  \centering
  \renewcommand{\arraystretch}{1.2}
  \begin{tabular}{llcc}
    \toprule
    \textbf{IPC} & \textbf{Model} & \textbf{Test Acc. (\%)} & \textbf{Oracle Acc. (\%)} \\
    \midrule
    \multirow{2}{*}{500} & Baseline ($r=1$) & $\mathbf{73.88}$ & $73.88$ \\
                         & MN ($r=32$) & $66.52$ & $\mathbf{99.12}$ \\
    \midrule
    \multirow{2}{*}{10}  & Baseline ($r=1$) & $12.94$ & $12.94$ \\
                         & MN ($r=32$) & $\mathbf{23.24}$ & $\mathbf{99.84}$ \\
    \bottomrule
  \end{tabular}
\end{table}

\subsubsection{Diversity utilization and path imbalance}

We next examine why this diversity is effective in low-data regimes but does not translate into equally large accuracy gains in data-rich regimes.
We interpret this difference from the viewpoint of path utilization, namely path imbalance.
If diverse paths are broadly utilized, their complementarity should be reflected in the final prediction.
Conversely, if learning becomes concentrated on only a subset of paths, the benefit of diversity may not be fully realized even when such diversity exists internally.

\begin{figure}
    \centering
    \includegraphics[width=1\linewidth]{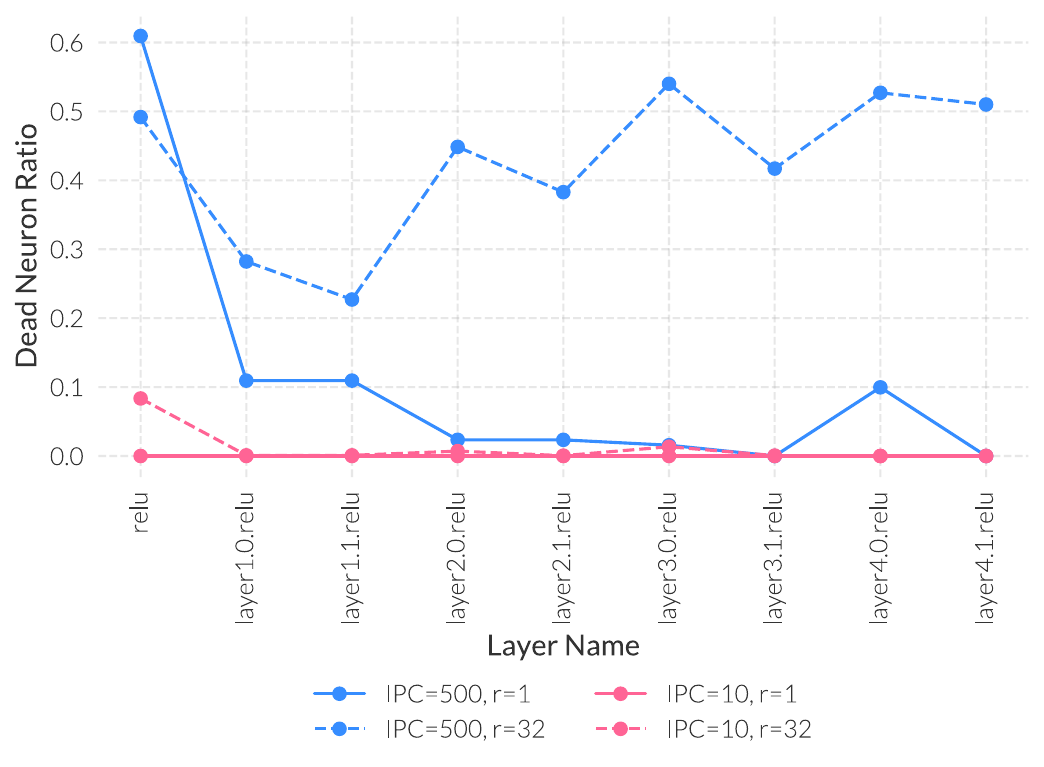}
    \caption{Layer-wise dead neuron ratio (DNR) under IPC=500 and IPC=10.
    In the data-rich regime, many channels in high-MN become effectively inactive in deeper layers.}
    \label{fig:dead_neuron_ratio}
\end{figure}

To assess the utilization efficiency of the feature space, we analyze the Dead Neuron Ratio (DNR).
A dead neuron refers to a unit that remains persistently inactive, as in the well-known dying ReLU problem~\cite{Lu_2020}.
To quantify this phenomenon, we define DNR as the proportion of channels whose output after ReLU activation is always zero for all samples in the validation set.
As shown in \figref{fig:dead_neuron_ratio}, DNR remains relatively low even in deeper layers under the low-data regime (IPC=10).
In contrast, under the data-rich regime (IPC=500), DNR increases sharply in the deeper layers of the high-MN model, indicating that many channels are effectively unused.
This suggests that, although high-MN models possess substantial internal diversity in data-rich regimes, only a subset of it is actually utilized.

\begin{figure*}[htbp]
    \centering
    \begin{minipage}[b]{0.48\textwidth}
        \centering
        \includegraphics[width=7.5cm]{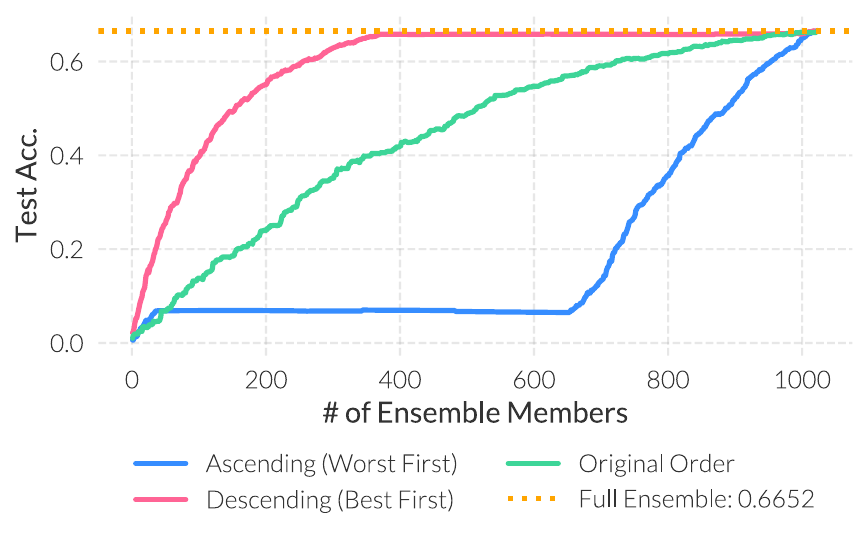}
        \text{(a) IPC=500}
    \end{minipage}
    \hfill
    \begin{minipage}[b]{0.48\textwidth}
        \centering
        \includegraphics[width=7.5cm]{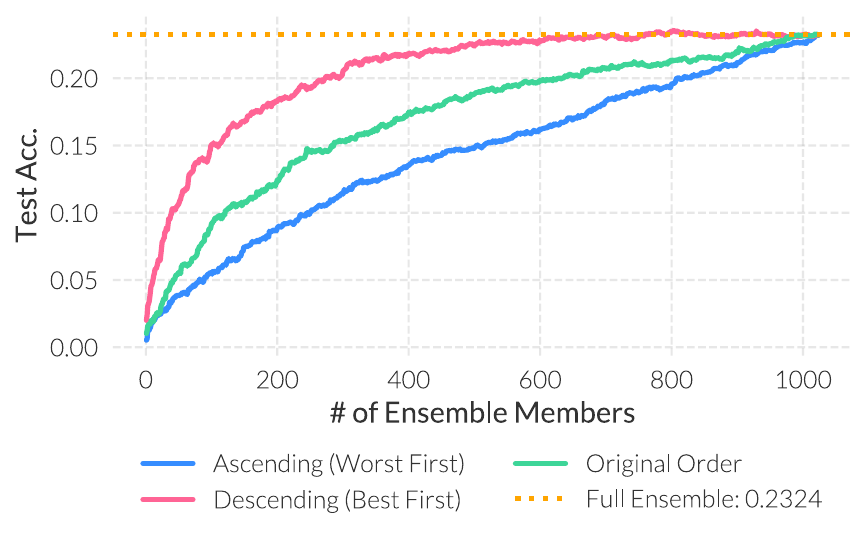}
        \text{(b) IPC=10}
    \end{minipage}
    \caption{Cumulative ensemble accuracy obtained by progressively adding paths.
    In the data-rich regime, prediction is dominated by a small subset of strong paths, whereas in the low-data regime, contributions are more broadly distributed.}
    \label{fig:cumulative_acc}
\end{figure*}

To further test this hypothesis, we sort the paths according to their individual accuracies and evaluate the cumulative ensemble accuracy obtained by progressively averaging them.
Specifically, paths are added in three orders:
(i) Best-first (descending individual accuracy),
(ii) Worst-first (ascending individual accuracy), and
(iii) Original (the original ordering).
As shown in \figref{fig:cumulative_acc}, when IPC=500, Best-first rises rapidly after only a few paths and reaches nearly the final performance early, whereas Worst-first shows almost no improvement in the initial stage.
This indicates that the final prediction is heavily dominated by a small subset of strong paths, while the remaining paths contribute very little in practice.
In contrast, when IPC=10, the gap between Best-first and Worst-first is relatively small, indicating that many paths make meaningful and complementary contributions to the final prediction.

Taken together, these results suggest that the advantage of high MN strength depends not merely on its ability to generate path-wise diversity, but on how broadly that diversity is utilized during learning.
In the low-data regime, many paths remain active and contribute complementarily to prediction, hence diversity directly translates into improved generalization.
In the data-rich regime, by contrast, diversity may still exist internally, but learning becomes concentrated on a subset of paths, preventing the full benefit of that diversity from being reflected in the aggregated prediction.
Therefore, the answer to RQ2 can be summarized as follows:
high-MN is advantageous in low-data regimes because it induces diverse path-wise representations and allows many of them to contribute complementarily, whereas in data-rich regimes the same diversity is only partially utilized due to path imbalance.

\subsection{Computational implications of Multi-Narrow transformation}
\label{sec:cost}

Finally, we examine the practical cost of the MN transformation.
Table~\ref{tab:cost} reports the MACs, forward time, and VRAM usage of ResNet-18 models with different MN strengths \(r\). All computational-cost measurements were conducted using a single NVIDIA RTX A6000 GPU and an AMD EPYC 7302 16-Core Processor. Forward latency and VRAM usage were measured with a batch size of 128 under the standard CIFAR-100 input resolution.

\begin{table}[tb]
    \centering
    \caption{Computational cost of MN models with different transformation strengths \(r\).
    MACs are reported for training (T) and evaluation (E), while forward time and VRAM usage are measured empirically.}
    \label{tab:cost}
    \resizebox{\linewidth}{!}{
    \begin{tabular}{c c c c c c c}
        \toprule
        \multirow{2}{*}{$r$} & \multicolumn{2}{c}{MACs} & \multicolumn{2}{c}{Forward [ms]} & \multicolumn{2}{c}{VRAM [MB]} \\
        \cmidrule(lr){2-3} \cmidrule(lr){4-5} \cmidrule(lr){6-7}
         & (T) & (E)  & (T) & (E)  & (T) & (E) \\
        \midrule
        1  & 71.5G & 71.3G & 5.0   & 4.2   & 381.8   & 160.0 \\
        2  & 72.1G & 71.6G & 8.4   & 7.3   & 680.5   & 261.7 \\
        4  & 73.4G & 72.4G & 15.3  & 13.2  & 1273.8  & 473.8 \\
        8  & 75.9G & 74.0G & 32.1  & 28.7  & 2540.9  & 928.9 \\
        16 & 80.9G & 77.1G & 62.7  & 56.5  & 5277.5  & 1986.7 \\
        32 & 90.9G & 83.4G & 132.8 & 121.1 & 11633.9 & 3655.8 \\
        \bottomrule
    \end{tabular}
    }
\end{table}

The increase in MACs is relatively moderate: even at \(r=32\), MACs are only about \(1.27\times\) the baseline (\(r=1\)) during training and \(1.17\times\) during evaluation.
In contrast, the measured forward time and VRAM usage increase sharply with \(r\).
At \(r=32\), the forward time reaches 132.8\,ms for training and 121.1\,ms for evaluation, corresponding to approximately \(26.6\times\) and \(28.8\times\) the baseline, respectively.
VRAM usage also rises to 11633.9\,MB during training and 3655.8\,MB during evaluation, corresponding to approximately \(30.5\times\) and \(22.8\times\) the baseline.

These results show that the cost of the MN transformation cannot be adequately described by theoretical computation alone.
Despite approximate budget preservation in dense-coupling layers, partitioning the model into many narrow paths introduces substantial practical overhead through increased activation storage and reduced hardware efficiency.
Therefore, although high MN strength can be an effective design choice in the low-data regime, its accuracy gains must be weighed carefully against its computational cost in practical deployment.

\section{Discussion}
\label{sec:discussion}
\subsection{Implications for capacity allocation}
\label{sec:discussion_implications}

Our results show that the preferable allocation of capacity in SMEs is not uniform, but depends strongly on the training-data regime.
When sufficient training data are available, SW or weakly transformed MN configurations achieve more stable and higher performance, whereas in the low-data regime, configurations with high MN strength become advantageous across multiple architectures and datasets.
This suggests that, under approximately fixed budget conditions, the choice between making a model wider and making it more multiplicative should depend on the data regime.

Our internal analyses further show that this trend is related to both the emergence of path-wise diversity and the extent to which that diversity is utilized.
High-MN configurations tend to form diverse and complementary path-wise representations, but this does not always translate into improved final accuracy.
In particular, in the data-rich regime, learning becomes concentrated on a subset of paths, and only part of the available diversity is effectively exploited.
Thus, what matters is not simply the presence of diversity itself, but how broadly that diversity is utilized during learning.
Although high MN strength can be an effective option in the low-data regime, it also incurs increased forward time and VRAM usage, implying that the trade-off between accuracy gain and computational efficiency must be carefully considered in practical deployment.

Beyond this accuracy--efficiency trade-off, the value of MN should not be reduced to predictive performance alone.
Because MN explicitly factorizes a model into path-wise independent components, it provides a structurally transparent way to control how capacity is distributed between width and multiplicity.
Our results also suggest that this factorization can induce diverse and complementary internal representations without relying on auxiliary collaborative losses, which makes MN a useful setting for analyzing ensemble behavior within a single model.
Moreover, such modular path structure may be attractive for future extensions, including path-wise pruning, selective execution, contribution-based diagnosis, and distributed deployment across heterogeneous resources.
Although the present implementation does not provide a general efficiency advantage in terms of latency or VRAM usage, the structural modularity of MN remains an important practical and conceptual merit.

\subsection{Scope and limitations}
\label{sec:discussion_scope}

This study focuses on CNN-based image classification models, for which differences in capacity allocation induced by the MN transformation can be compared relatively directly, and our experiments are conducted primarily in the scratch-training setting.
Accordingly, our conclusions should be understood within this scope, and extending them to Transformer-based models or transfer-learning settings remains an important direction for future work.
In such settings, self-attention mechanisms and pretrained representations may strongly influence performance, and therefore the trends observed in this study may not hold in the same form.

In addition, we deliberately avoided per-path supervision and auxiliary collaborative losses in order to isolate the effect of structural capacity allocation itself.
This design choice helped reveal the failure mode of path imbalance in the data-rich regime, but it also leaves room for future learning strategies that may mitigate this issue.
Furthermore, because our hyperparameter settings do not constitute a fully exhaustive best-to-best tuning for all conditions, the precise location of the transition boundary may vary quantitatively.
Even so, the central conclusion of this study lies not in a single numerical threshold, but in the consistent trend that the preferable capacity allocation changes with the training-data regime.

\section{Conclusion}
\label{sec:conclusion}

In this study, we examined whether a single-wide or a multi-narrow configuration is preferable under approximately matched parameter budgets by systematically varying the amount of training data.
Our results show that the effectiveness of the MN transformation depends strongly on the training-data regime: when sufficient data are available, SW or weakly transformed MN configurations are preferable, whereas in the low-data regime, high MN strength provides a consistent advantage.
We further showed that this advantage is associated with increased path-wise diversity and the complementary use of multiple paths, while in the data-rich regime its benefit is not fully realized because of path imbalance.
These findings suggest that the preferable design of capacity allocation in SMEs is not fixed, but should be chosen according to the data regime.

\appendix
\section{Detailed results across datasets and data regimes}

\begin{figure*}[htbp]
    \centering
    \begin{minipage}[b]{0.39\textwidth}
        \centering
        \includegraphics[width=7cm]{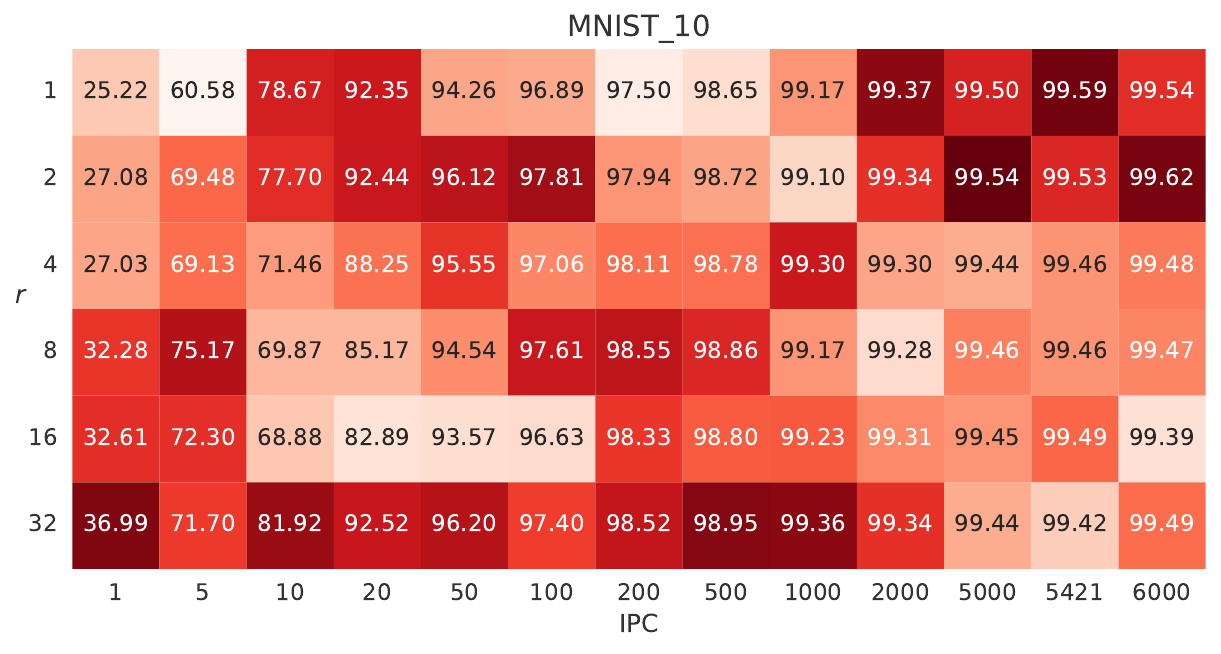}
    \end{minipage}
    \hfill
    \begin{minipage}[b]{0.34\textwidth}
        \centering
        \includegraphics[width=6.5cm]{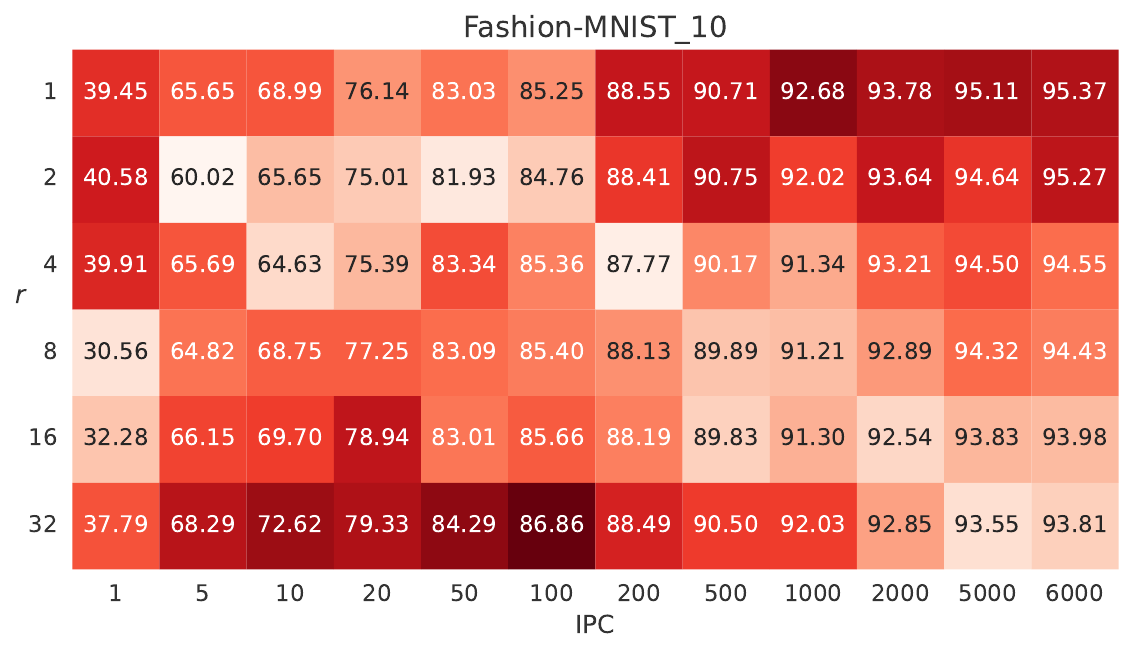}
    \end{minipage}
    \hfill
    \begin{minipage}[b]{0.25\textwidth}
        \centering
        \includegraphics[width=3.5cm]{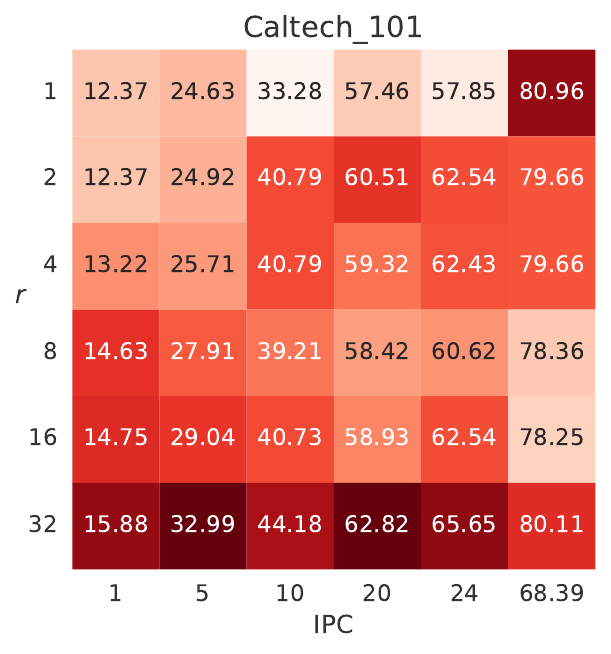}
    \end{minipage}
    \hfill
    \vspace{0.1cm}
    
    \begin{minipage}[b]{0.37\textwidth}
        \centering
        \includegraphics[width=6.5cm]{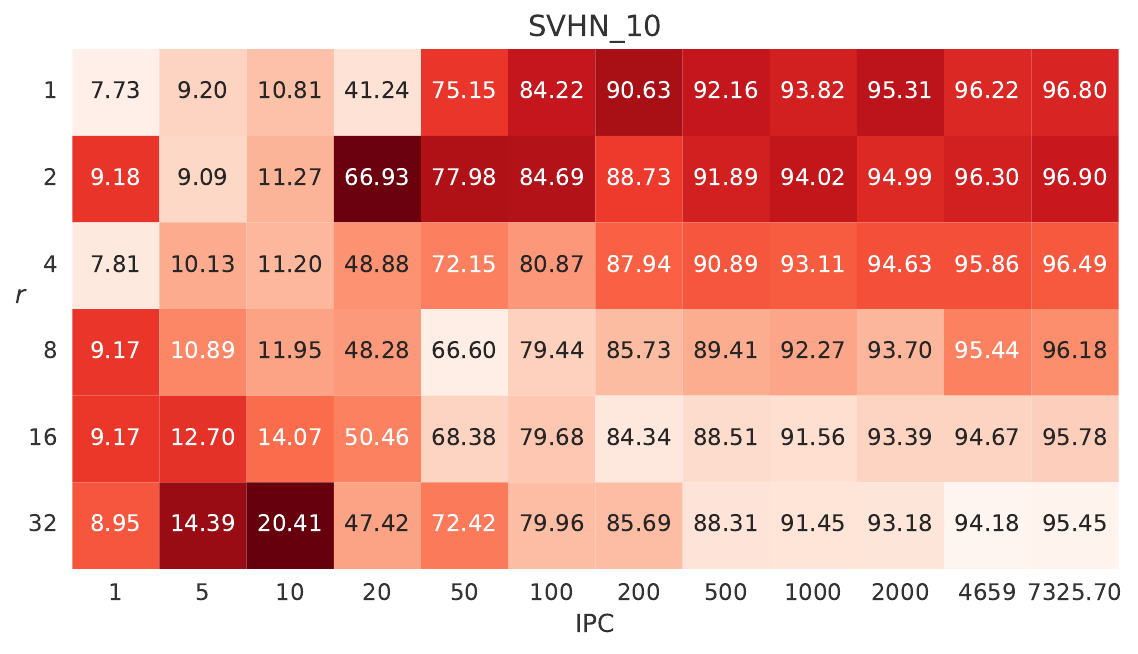}
    \end{minipage}
    \hfill
    \begin{minipage}[b]{0.33\textwidth}
        \centering
        \includegraphics[width=6.cm]{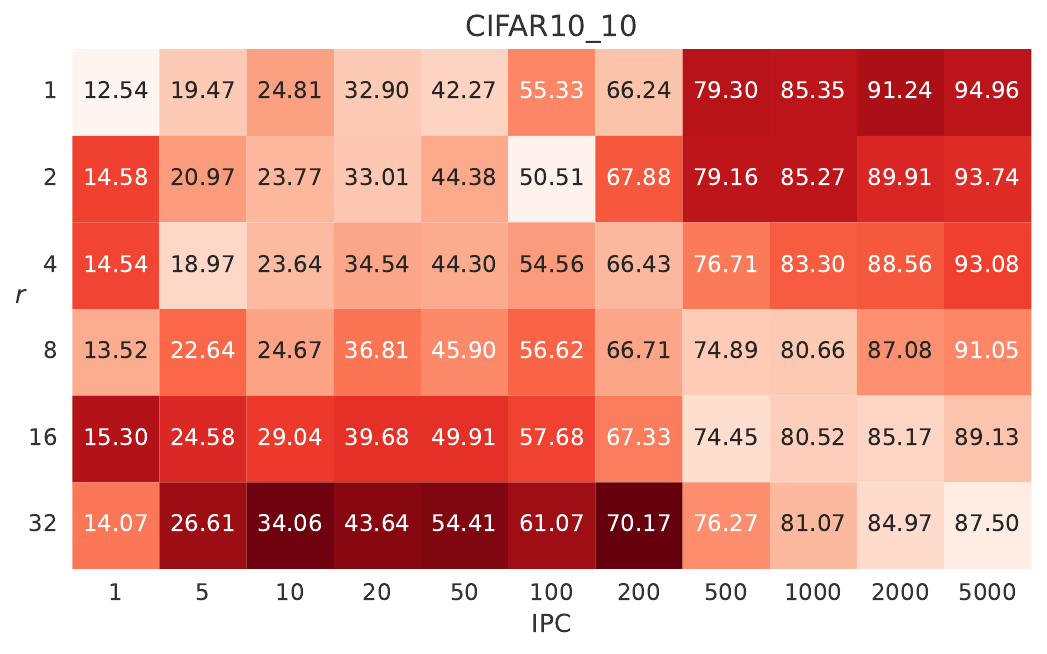}
    \end{minipage}
    \hfill
    \begin{minipage}[b]{0.28\textwidth}
        \centering
        \includegraphics[width=4.5cm]{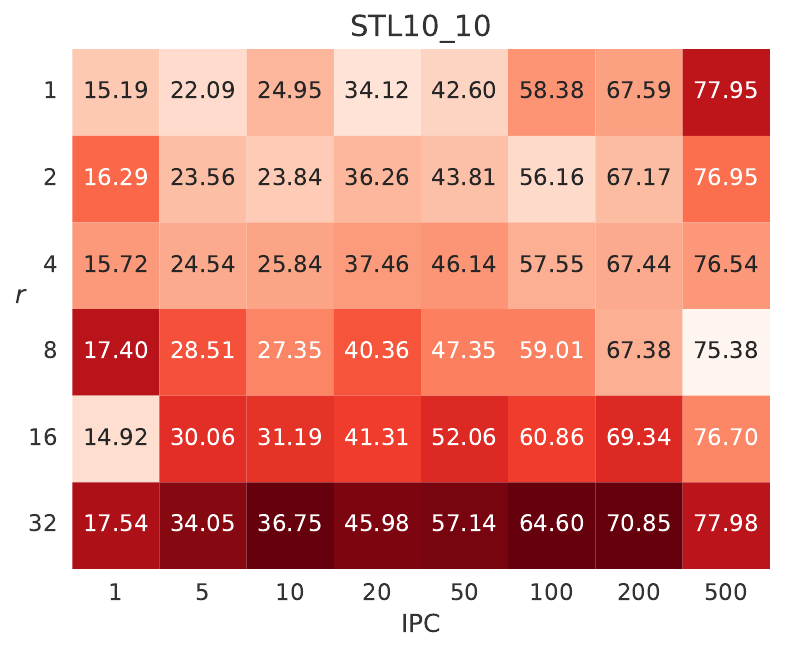}
    \end{minipage}
    \hfill
    \vspace{0.1cm}

    \begin{minipage}[b]{0.27\textwidth}
        \centering
        \includegraphics[width=4.8cm]{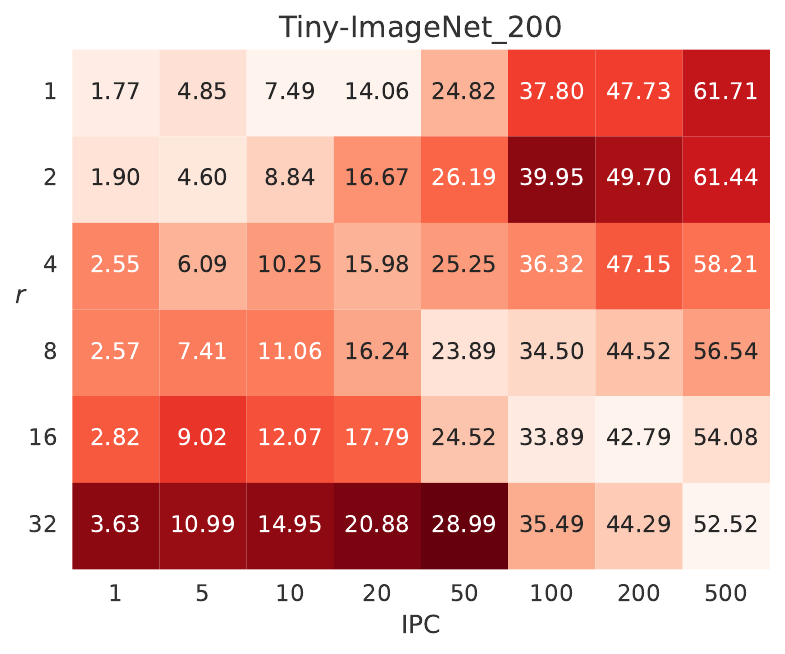}
    \end{minipage}
    \hfill
    \begin{minipage}[b]{0.15\textwidth}
        \centering
        \includegraphics[width=2.15cm]{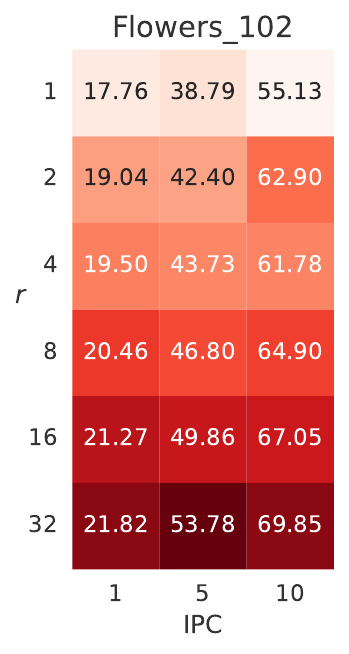}
    \end{minipage}
    \hfill
    \begin{minipage}[b]{0.24\textwidth}
        \centering
        \includegraphics[width=3.8cm]{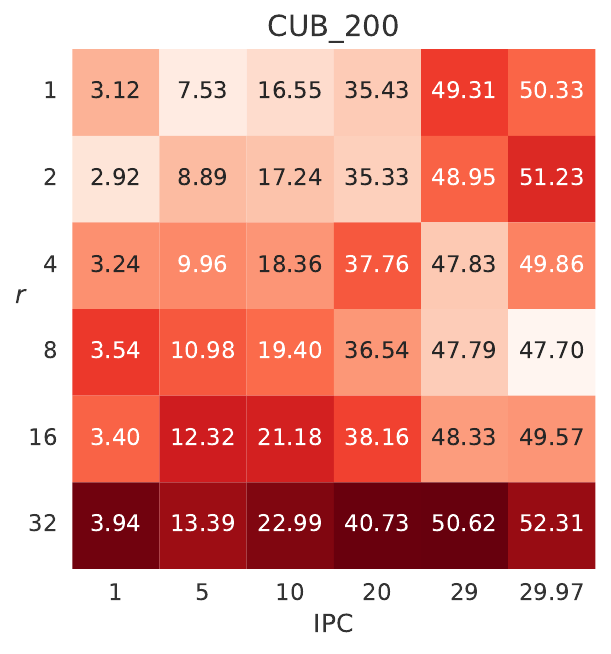}
    \end{minipage}
    \hfill
    \begin{minipage}[b]{0.32\textwidth}
        \centering
        \includegraphics[width=4.3cm]{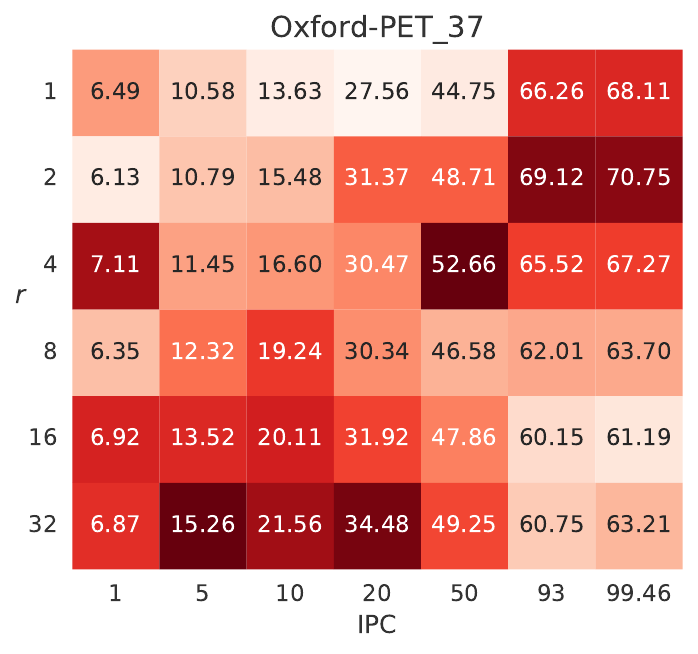}
    \end{minipage}

    \caption{Effect of MN strength and training-data size across datasets (ResNet-18).}
    \label{fig:ten_pdf}
\end{figure*}

This appendix reports the full IPC--MN-strength performance maps for all additional datasets included in the dataset-robustness analysis. These figures provide the detailed results underlying the bubble-chart summary shown in \secref{sec:rq2}, and the corresponding visualizations are presented in \figref{fig:ten_pdf}. Across most datasets, the high-MN model (\(r=32\)) tends to outperform the baseline model (\(r=1\)) in low-IPC regimes. These figures complement the summary visualization in the main text and further support the view that this tendency is broadly consistent across datasets.

\section{Parameter-count changes across architectures}
\label{app:param_variation}

Table~\ref{tab:param_gain_arch} reports the parameter counts of each architecture for different MN strengths \(r\), along with their increase relative to the baseline configuration (\(r=1\)). Although the Multi-Narrow (MN) transformation is intended to approximately preserve the parameter budget, the parameter count is not strictly constant for all architectures.

For ResNet-18 and Wide-ResNet-50, the increase remains small even at large \(r\), indicating that the transformation is nearly parameter-preserving in these architectures. By contrast, MobileNetV2, EfficientNet-B0, and RegNetY-400MF show substantial parameter inflation as \(r\) increases, while ConvNeXt-Tiny exhibits a moderate increase. This difference arises from architecture-dependent operator design, and should be taken into account when interpreting cross-architecture SW/MN comparisons.
\begin{table}[t]
\centering
\caption{Parameter counts (M) after Multi-Narrow transformation. The second row for each architecture shows the increase relative to \(r=1\).}
\label{tab:param_gain_arch}
\resizebox{\linewidth}{!}{%
\begin{tabular}{llcccccc}
\toprule
Architecture & Metric & \(r=1\) & \(r=2\) & \(r=4\) & \(r=8\) & \(r=16\) & \(r=32\) \\
\midrule
\multirow{2}{*}{ResNet-18}
  & Params [M] & 11.2 & 11.3 & 11.4 & 11.7 & 12.2 & 13.3 \\
  & Gain [\%]  & +0.0 & +0.9 & +1.8 & +4.5 & +8.9 & +18.8 \\
\midrule
\multirow{2}{*}{WResNet-50}
  & Params [M] & 67.0 & 67.3 & 67.9 & 69.0 & 71.2 & 75.7 \\
  & Gain [\%]  & +0.0 & +0.4 & +1.3 & +3.0 & +6.3 & +13.0 \\
\midrule
\multirow{2}{*}{MobileNetV2}
  & Params [M] & 2.4 & 2.6 & 3.0 & 3.9 & 5.8 & 9.5 \\
  & Gain [\%]  & +0.0 & +8.3 & +25.0 & +62.5 & +141.7 & +295.8 \\
\midrule
\multirow{2}{*}{EfficientNet-B0}
  & Params [M] & 4.1 & 4.5 & 5.2 & 6.7 & 9.6 & 15.7 \\
  & Gain [\%]  & +0.0 & +9.8 & +26.8 & +63.4 & +134.1 & +282.9 \\
\midrule
\multirow{2}{*}{RegNetY-400MF}
  & Params [M] & 3.9 & 4.0 & 4.2 & 4.5 & 9.9 & 23.3 \\
  & Gain [\%]  & +0.0 & +2.6 & +7.7 & +15.4 & +153.8 & +497.4 \\
\midrule
\multirow{2}{*}{ConvNeXt-Tiny}
  & Params [M] & 27.9 & 28.4 & 29.3 & 31.2 & 34.9 & 42.5 \\
  & Gain [\%]  & +0.0 & +1.8 & +5.0 & +11.8 & +25.1 & +52.3 \\
\bottomrule
\end{tabular}%
}
\end{table}

\printcredits

\bibliographystyle{cas-model2-names}

\bibliography{mybib2}

\bio{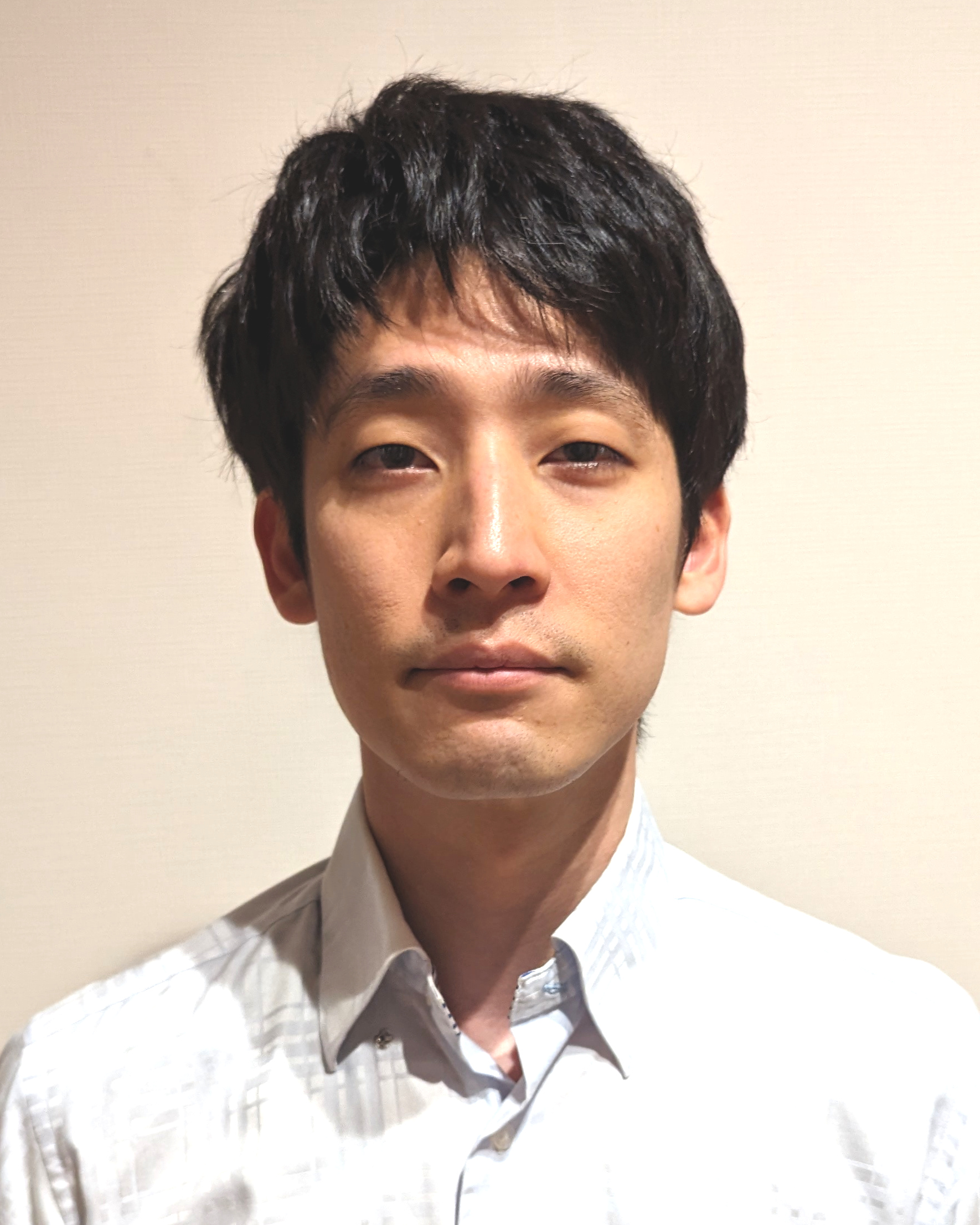}
\textbf{Tatsuhito Hasegawa} received the Ph.D. degree in engineering from Kanazawa University, Kanazawa, in 2015. From 2011 to 2013, he was a System Engineer with Fujitsu Hokuriku Systems Ltd. From 2014 to 2017, he was an Assistant with Tokyo Healthcare University. From 2017 to 2020, he was a Senior Lecturer with the Graduate School of Engineering, University of Fukui. He is currently an Associate Professor. His study interests include human activity recognition, deep learning, and intelligent learning support system. He is also a member of IPSJ and JSAI.
\endbio

\bio{}
\textbf{Taisei Tanaka} received the B.S. degree in engineering from the University of Fukui, Japan, in 2024, and the M.S. degree from the Graduate School of Engineering, University of Fukui, Japan, in 2026. He has since completed his graduate studies and is currently employed as an engineer. His research interests include deep ensemble learning.
\endbio

\end{document}